\documentclass[letterpaper]{article} 
\usepackage{aaai18}  
\usepackage{times, amsfonts, amsmath}  
\usepackage{helvet}  
\usepackage{courier}  
\usepackage{url}  
\usepackage{graphicx,, algorithm, algorithmic}  
\usepackage{hyperref, todonotes}

\usepackage{xcolor}
\usepackage{tikz}
\usetikzlibrary{er}
\tikzstyle{every entity} = [draw = orange , fill = orange!20]
\tikzstyle{every attribute} = [ draw = purple , fill = purple!20]
\tikzstyle{every relationship} =  [draw = teal , fill = teal!20]
\tikzstyle{derived attribute} = [ attribute , dashed]
\tikzstyle{multi attribute} = [  attribute , double distance =1.5 pt]
\tikzstyle{total} = [ double distance =1.5 pt]

\newcommand { \key }[1]{ \underline {#1}}

\title{Relational Constraints for Metric Learning on Relational Data}
\author{Jiajun Pan, Hoel Le Capitaine \& Philippe Leray\\
LS2N, UMR CNRS 6004\\
University of Nantes\\
44300 Nantes Cedex, France \\
\texttt{\{jiajun.pan, hoel.lecapitaine, philippe.leray\}@ls2n.fr} \\
}

\hypersetup{draft}
\begin{document}

\maketitle

\begin{abstract}
Most of metric learning approaches are dedicated to be applied on data described by feature vectors, with some notable exceptions such as times series, trees or graphs. The objective of this paper is to propose a metric learning algorithm that specifically considers relational data. The proposed approach can take benefit from both the topological structure of the data and supervised labels. For selecting relative constraints representing the relational information, we introduce a link-strength function that measures the strength of relationship links between entities by the side-information of their common parents. We show the performance of the proposed method with two different classical metric learning algorithms, which are ITML (Information Theoretic Metric Learning) and LSML (Least Squares Metric Learning), and test on several real-world datasets. Experimental results show that using relational information improves the quality of the learned metric.
\end{abstract}

\section{Introduction}
\label{sec:int}

Sample similarity measurement lies at the heart of many classification and clustering methods in pattern recognition and machine learning. For instance, in classification, the k-Nearest Neighbor classifier uses a metric to identify the nearest neighbors; in clustering algorithms, k-means rely on distance measurements between data points; in information retrieval, documents are often ranked according to their relevance to a given query based on similarity scores. The performance of these algorithms rely on the quality of the metric. The conventionally used Euclidean distance cannot give a convenient dissimilarity in many cases, due to the distribution of the data (see \cite{tenenbaum2000global}). Thus, it calls a great need for appropriate ways to measure the distance or similarity between observations in  learning algorithms.
 
Metric learning has now been used for more than a decade to deal with this problem, and can be seen a feature/representation learning  allowing the use of Euclidean distances later on. The vast majority of metric learning approaches are dedicated to be applied on data described by feature vectors, where the objective is generally to learn a matrix $M$ that is used for the Mahalanobis distance \[D_M^2(\mathbf{x},\mathbf{y}) = (\mathbf{x}-\mathbf{y})^T M (\mathbf{x}-\mathbf{y}),\] 
where $\mathbf{x}$ and $\mathbf{y}$ are $d$-dimensional vectors and $M$ is a positive semi-definite matrix. Using such a metric is equivalent to  perform a linear projection of the data by the matrix decomposition $M=L^TL$, where $L \in \mathbb{R}^{k \times d}$, and then use the conventional Euclidean distance in this new feature space.

The interested reader can refer to reviews in \cite{kulis2013metric} and \cite{bellet2015metric}. 

In this model, $M$ contains the learned parameters and is learned mostly from  supervised information. Most of the approaches make use of label constraints, which means that the constraints are generated by the target labels or other supervised information. Such a distance is perfectly adapted for flat or iid data, but obviously fails to take into account complex and/or (semi-)structured, non-iid data without considering the structured information.
 
There are some notable exceptions such as times series in  \cite{garreau2014metric} (through dynamic time warping methods) and trees or graphs in \cite{bellet2016learning}  (by using an edit distance) or networks, proposed in \cite{shaw2011learning}.
Relational databases are increasingly used in almost all applications. A lot of real world datasets present aspects of multi-relation between the observations. For instance, social service networks, Wikipedia network, molecular biology classification and so forth. These databases are organized based on a relational model of data which contains entity tables and association tables between entity tables. Using this data in machine learning is now under consideration for years \cite{getoor2007introduction}, but to the best of our knowledge, no attention on metric learning has been paid for such data. Naturally, one can use traditional metric learning algorithms for individual entities, but at the price of losing rich information coming from the relational structure of the data. Taking good use of associations between entities can help to improve metric performance.  
 
The goal of this paper is to propose the use of both relational information and supervised information in constraints selection for metric learning. Such a definition allows to build rich models, which can eventually be used for domain adaptation, transfer learning, feature learning and data visualization with both flat and multi-relational data. In particular, we propose a solution that is able to incorporate relational information within metric learning, and then illustrate its benefit compared to traditional flat approaches.

Note that the proposed approach starts from (hyper)graph data, where as approaches as in \cite{dhillon2012metric}. 
Instead of using label constraints, the authors used graph based constraints which enforces the distance between the unlinked node and target node to be bigger than the distance between the k-farthest linked node and target node. 
We propose a more general model denoted as link-strength constraints which is generated by a link-strength function measuring the similarity of nodes by the side information of the relationships between them. The proposed link-strength function gives the possibility of using similarity learning  to encode the relational information into the constraints for metric learning algorithm. 
 
This paper is organized as follows. We recall basic definitions related  to metric learning, as well as related works, in section \ref{sec:rel}. In section \ref{sec:pro}, we present our approach, which is validated by experimental comparisons on real-world datasets in section \ref{sec:exp}. Final comments and perspectives are drawn in section \ref{sec:con}. 

\section{Related Works}
\label{sec:rel}

\subsection{Metric Learning and Relational Learning}

Since its seminal paper in 2003 \cite{xing2003distance},  there have been many propositions dealing with learning metrics. In \cite{kulis2013metric}, one can find a number of different metric learning algorithms. Most of the propositions of metric learning rely on a new metric $D'(\mathbf{x}, \mathbf{y})$ as $D(f(\mathbf{x}), f(\mathbf{y}))$ with a mapping function $f$. The metric learning then simply consists in learning the projection $f$ by using constraints on the (dis)similarity on observations. With constraints generated from given information, metric learning approaches generally consider the generic loss function written as
\[ L(M) = \sum_{(i,j,k) \in \mathcal{C}} \ell_M(i,j,k) +\lambda r(M),\]
where $\ell_M(i,j,k)$ is the encoded loss from 
pairs of entity nodes (i,j)  in set $\mathcal{S}=\{(i,j)\}$ of similar observations and pairs of entity nodes (i,k)  in set $\mathcal{D}=\{(i,k)\}$ of the dissimilar observations. $r(M)$ is a regularization term on the matrix $M$ (e.g Frobenius norm, trace-norm). 
The loss $ \mathcal{C} \ell_M(i,j,k) $ is then generally written as the hinge loss functions $max(0,D_M(\mathbf{x}_i,\mathbf{x}_j) - u)$ and $ max(0,l-D_M(\mathbf{x}_i,\mathbf{x}_k))$ with threshold parameters $u$ and $l$ or $ max(0,m+D_M(\mathbf{x}_i,\mathbf{x}_j)-D_M(\mathbf{x}_i,\mathbf{x}_k))$ with margin parameter $m$. 

Traditionally, in metric learning, one uses label constraints in order to select  similar and dissimilar constraint sets by the target label or other supervised information. In this case, the pair $(i,j)$ contains the nodes in same class, while the labels of nodes $(i,k)$ are different. 

Relational learning deals with learning models for which data consists in a generally complex relational structure. As a difference with flat datasets, the main learning tasks pay more attention on supervised information from the relations and knowledge from the topology of the relational graph. Collective classification is classification of related entities that may share identical classes \cite{sen2008collective} from the relationship information.

Note that, in \cite{dumanvcic2017demystifying}, the authors propose an analysis of the meaning of the latent space learned by a deep learning algorithm on relational datasets, mainly because of the black box problem of deep approaches. The same analysis can be conducted for the meaning of metric learning, which  can also  be treated as mapping original feature space to a latent space. This paper mainly focuses on the usefulness of latent space and the redundancy of the latent features. They show the good performance of using unsupervised relational information for a classifier in a latent space. We specifically focus on using  relational information for collective classification. 

For collective classification, there are several approaches to learn metric with relational information. In \cite{kramer2001propositionalization}, they use graph relational information with propositionalization, transfer the relational representation of a learning problem into a propositional representation. Another approach uses metric learning on graphs for domain adaptation \cite{dhillon2012metric}.  To this aim, they propose an iterative learning algorithm on the graph. From the resource domain, the nodes with labels, they  learn a new metric and apply it on related target domain, the nodes without labels. Then, the graph is updated depending on the learned distance, and constraints with low entropy instances are selected for next iteration. 

Those approaches do not consider the different modalities of the observations (features and relations). In many cases, the structure of the data does not allow to directly measure distances as if the observations were belonging to an Euclidean space. In particular, complex and structured data needs to be processed in a different way than usual tabular data. Such complex data includes times series, videos, graphs, relational data, without exhaustivity.

Some metric learning algorithms consider relational data as heterogeneous networks for each different relationships. For instance, in \cite{zhai2013heterogeneous}, they propose a  heterogeneous metric learning algorithm, which  integrates the structure of different graphs into a joint graph regularization. They use two mapping function for the feature space of the object entities and subject entities in one relation, and then introduce a joint graph regularization for iterative optimize the loss function. In \cite{dong2017metapath2vec}, they start from the same principle but use  meta-path-based random walks to incorporate the heterogeneous network structures into skip-gram vectors for dealing with the relational graph. Those algorithms use joint regularization for different entities in heterogeneous networks, with good performance on considering the structure information in relational dataset. However, without considering the side information in the relational links. It  processes the relational variables the same way as the entities and subject to their algorithm, but it ignores the differences between entity tables and association tables. Our proposed method includes the value of different variables on the relationship in the datasets and distinguish them with entities. 

In this paper, we consider the case of relational data, where several tabular datasets (entities) are linked together through associations. The basic principle of our approach is to use relational links between entities when setting the constraints of the metric learning algorithm. Consequently, this approach can be used in any constraint-selection based metric learning algorithm.

\subsection{Metric Learning with Relational Constraints}

Recently, some metric learning approaches have been focusing on graph data. In this context, the structural information consists in the presence or absence of links between nodes of the graph. Constructing the set of similar nodes and dissimilar nodes then just uses the adjacency matrix of the graph.
For example, the simplest relative link constraints is $D_M(\mathbf{x}_i,\mathbf{x}_j)\le D_M(\mathbf{x}_i,\mathbf{x}_k)+m,(i,j) \in \mathcal{S},(i,k) \in \mathcal{D}$ with the adjacent matrix $A_{ij}=1,A_{ik}=0,\forall (i,j)\in \mathcal{S},\forall (i,k) \in \mathcal{D}$, which only check the relative distance between the connected and disconnected relational links. In the sequel, this method is termed as relative link constraint.

In \cite{shaw2011learning}, they formulate the metric in a preserving embedding structure and learn from linear constraints with the graph topology. They proposed nearest neighbor graphs and maximum weight subgraphs, which are two ways for generating the supervised constraints with the relational information in graphs. The nearest neighbour constraints is denoted as $D_M^2(\mathbf{x}_i,\mathbf{x}_j) > (1-A_{ij})\max_l(A_{il}D_M^2(\mathbf{x}_i,\mathbf{x}_l)),\forall i,j$ which aims to constraint the disconnected node to be more far from the target node than its farthest connected node (neighbour). 

However, this approach does not take into account  that  there are a lot of information in the valued link with numerical or categorical variables. Furthermore, target labels of nodes are not used in this model. We propose to define new relative link constraints from to the link-strength constraints, with a link-strength function to measure the importance of the relational links between entities, as well as using the supervised information obtained from the labels.

\section{On selecting constraints with link-strength function}
\label{sec:pro}

In this paper, our objective is to propose a new approach of metric learning considering enforcing constraints with both the relational links and label information. The basic statement behind our proposal is that we consider that two connected individuals are more similar than two unconnected individuals. We then evaluate the amount of link similarity between individuals by considering their common parents in the graph. If, furthermore, labels are available for individuals, this can be incorporated into the learning algorithm. Consequently, our approach can be both supervised or unsupervised, depending on the availability of the information in the data.

We extend the relative link constraints by separating $\mathcal{S}$ and $\mathcal{D}$ with a dedicated link-strength function. The proposed constraints enhance the classical metric learning algorithms using relative constraints, such as ITML \cite{davis2007information} and LSML \cite{liu2012metric}.


A link-strength function $LS(\mathbf{x}_i,\mathbf{x}_j | r)$ is a function with the input is the relational information between the entities nodes $\mathbf{x}_i$ and $\mathbf{x}_j$ and the output is a real value to measure the "strength" or the similarity of the two nodes. The symbol $r$ represents the relationship information of the entities nodes $\mathbf{x}\in R^{n\times d}$, where $n$ is the number of instances and $d$ the number of node attributes. 

There are many ways to encode the relational information for this link-strength function, and we choose side-information of the common parents between the input nodes.

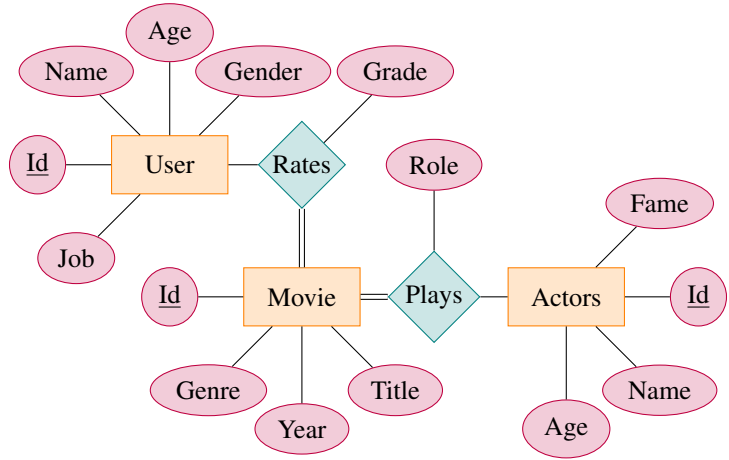
\begin{figure*}
\centering
\begin{tikzpicture}[node distance=5em]
\node[entity] (person) {User};
\node[attribute] (pid) [left of=person] {\key{Id}} edge (person);
\node[attribute] (name) [ above left of = person ] { Name } edge (person);
\node[attribute] ( phone ) [ above of = person ] { Age } edge (person);
\node[attribute] (address) [ above right of = person ] { Gender } edge (person);
\node[attribute] (address) [  below left of = person ] { Job } edge (person);
\node[relationship] (uses) [ right of = person ] { Rates } edge (person);
\node[attribute] (Grade) [above right of = uses] {Grade} edge (uses);
\node[entity] (tool) [ below of = uses ] { Movie } edge [ total ] (uses);
\node[attribute] (tid) [ left of = tool ] { \key{Id}} edge (tool);
\node[attribute] (tname) [ below right of = tool ] { Title } edge (tool);
\node[attribute] (tage) [ below of = tool ] { Year } edge (tool);
\node[attribute] (tgender) [ below left of = tool ] { Genre } edge (tool);
\node[relationship] (plays) [right of = tool] {Plays} edge[total] (tool);
\node[entity] (actor) [right of = plays] {Actors} edge (plays);
\node[attribute] (Role) [above of = plays] {Role} edge (plays);
\node[attribute] (aid) [ right of = actor ] { \key{Id}} edge (actor);
\node[attribute] (aname) [ below right of = actor ] { Name } edge (actor);
\node[attribute] (aage) [ below of = actor ] { Age } edge (actor);
\node[attribute] (afame) [ above right of = actor ] { Fame } edge (actor);
\end{tikzpicture}
\caption{Typical example of an Entity-Association relational model}
\label{fig:rel}
\end{figure*}

A relational schema $R = R_e \cup R_r$ contains a set of relational information where $R_e$ denotes the set of groups and types between entities in same tables and $R_r$ denotes the set of reference links between different tables. For a relation subset $r_k \subseteq R_r$ including all references between two entity tables, for example like the reference Ratings between User and Movie  as shown in Figure \ref{fig:rel}, we consider it as a many-to-many relationship. Let $\mathit{P}_{ij}$ be the set of common parents of $\mathbf{x}_i$ and $\mathbf{x}_j$, and $\ell_{ij} = |\mathit{P}_{ij}|$, the number of common parents.  

\begin{figure}
\centering
\def\layersep{2.25cm}
\begin{tikzpicture}[shorten >=1pt,->,draw=black!50, node distance=\layersep, scale=0.8]
    \tikzstyle{every pin edge}=[<-,shorten <=1pt]
    \tikzstyle{neuron}=[circle,fill=black!25,minimum size=17pt,inner sep=0pt]
    \tikzstyle{input neuron}=[neuron, fill=orange!40];
    \tikzstyle{output neuron}=[neuron, fill=orange!30];
    \tikzstyle{hidden neuron}=[neuron, fill=orange!20];
    \tikzstyle{annot} = [text width=4em, text centered]

    \foreach \name / \y in {1,...,4}
        \node[input neuron] (I-\name) at (0,-\y) {$\mathbf{x}^1_\name$};

    \foreach \name / \y in {1,...,4}
        \path
            node[hidden neuron] (H-\name) at (\layersep,-\y cm) {$\mathbf{x}^2_\name$};

            \path (I-1) edge (H-1);
            \path (I-1) edge (H-2);
            \path (I-2) edge (H-1);
            \path (I-2) edge (H-2);
            \path (I-2) edge (H-3);
            \path (I-2) edge (H-4);
            \path (I-3) edge (H-2);
            \path (I-3) edge (H-3);
            \path (I-3) edge (H-4);
            \path (I-4) edge (H-2);
            \path (I-4) edge (H-3);


    \node[annot,above of=H-1, node distance=1cm] (hl) {Relation $R^2$};
    \node[annot,left of=hl] {Relation $R^1$};
\end{tikzpicture}
\caption{Bipartite relational graph for a \textit{many-to-many} relationship table. The common parents of $\{\mathbf{x}^2_2,\mathbf{x}^2_3\}$ is the set of entities $\mathit{P}_{\mathbf{x}^2_2,\mathbf{x}^2_3}=\{\mathbf{x}^1_2, \mathbf{x}^1_3, \mathbf{x}^1_4\}$}
\label{fig:cp}
\end{figure}
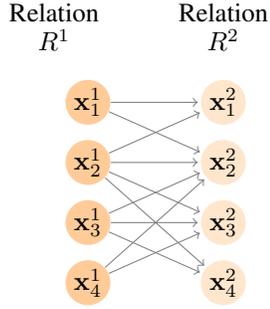

In Figure \ref{fig:cp}, we give a subsample of a bipartite relational graph, along with an example of common parents.

Naturally, the similar node would get the similar references from the same parents node. Consequently, we consider that the link-strength depends on the side-information of the common parents which are the values of the references.

\begin{figure*}
\centering
\includegraphics[width = \textwidth]{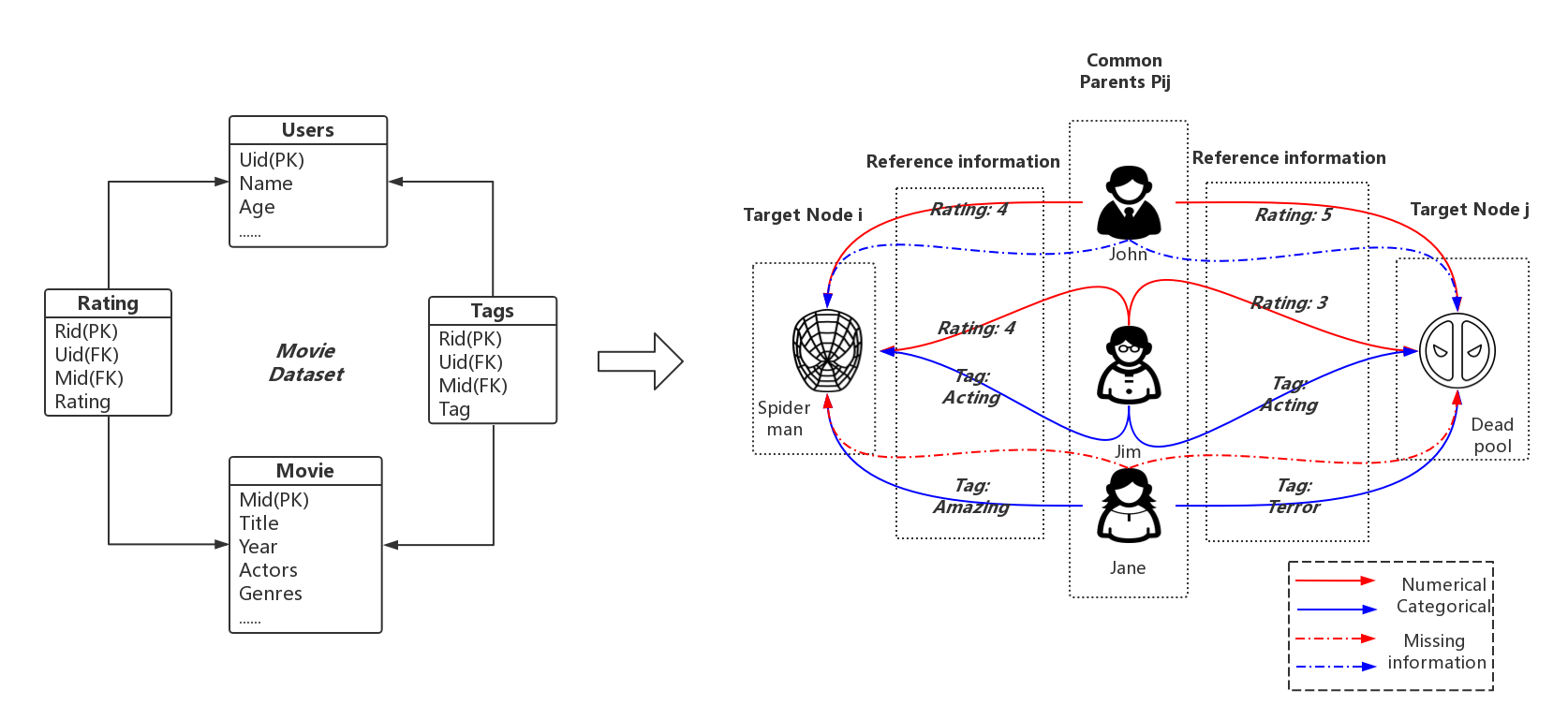}
\caption{The side information of references from common parents $P_{ij}$ to node $\mathbf{x}_i$  and node $\mathbf{x}_j$, in the movie dataset.}
\label{fig:cp2}
\end{figure*}

Additionally, the references can be quantified by $\alpha$ numerical variables $v$ and $\beta$ categorical variables $v^*$, see Figure \ref{fig:cp2}.


Given a relation $r_k$, we propose to defined the link-strength function as:

\begin{equation}
\label{eq:LS}
LS(\mathbf{x}_i,\mathbf{x}_j | r_k) =LS(\mathbf{x}_i,\mathbf{x}_j | \mathit{P}_{ij})$$ 
 $$ =  \sum_{h=1}^{\ell_{ij}} \left(\gamma \cdot w(h,i,j)+ (1-\gamma)  \cdot z(h,i,j) \right)
\end{equation}

where \[w(h,i,j) = \sum_{m=1}^\alpha  \exp\left(-\left|v_m(p_h,\mathbf{x}_i)-v_m(p_h,\mathbf{x}_j) \right|\right),\]  and  \[z(h,i,j) = \sum_{m=1}^{\beta} \left(v^*_m(p_h,\mathbf{x}_i) \circ v^*_m(p_h,\mathbf{x}_j) \right),\]
in which $x \circ y =1$ iff $x=y$, and $0$ otherwise, and $p_h$ is the $h$-th parent node in the set $\mathit{P}_{ij}$.
Note that numerical association attributes $v$ are normalized in the unit interval prior to link strength computation. Note also that we restrict to unit-length slot chains, i.e. the length of the sequence of foreign key references is equal to 1.
Then, we select the strongest links as similarity constraints, and the weakest links as dissimilarity constraints. The corresponding algorithm is given in Algorithm \ref{alg}. Remark that if two entities $\mathbf{x}_i$ and $\mathbf{x}_j$ do not have common parents, their link strength is zero, and therefore considered as dissimilar. 
\begin{algorithm}
\caption{Relational constraints learning \label{alg}}
\label{alg1}
\begin{algorithmic}[1]
\REQUIRE $N_{max}$ : number of desired constraints
\STATE $p \gets 1$ ;  $\mathcal{S} \gets \emptyset$ ; $\mathcal{D} \gets \emptyset$ 
\WHILE {$p \leq N_{max}$}
\STATE $X_p \gets (\mathbf{x}_i,\mathbf{x}_j) $ random pair generation
\STATE compute link strength $LS_p$ of $X_p$ using Equation (\ref{eq:LS})
\STATE $p \gets p +1$
\ENDWHILE
\WHILE {$|LS| > 0$}
\STATE $\mathcal{S} \gets \mathcal{S} \cup X_{\textrm{argmax} \{LS\}}$ 
\STATE $\mathcal{D} \gets \mathcal{D} \cup X_{\textrm{argmin} \{LS\}}$ 
\STATE $LS \gets LS \setminus \{\mathcal{S} \cup \mathcal{D}\}$
\ENDWHILE
\RETURN $\{\mathcal{S},\mathcal{D}\}$
\end{algorithmic}
\end{algorithm}

In this paper, we mainly focus testing the proposed algorithm on one $R_r$ relation dataset. However, this link-strength function could be easily extended to multi-relational datasets by summing the link-strength for each relations. It could also be extended from the reference relation $R_r$ to $R_e$, that consider the group structures as every edge between nodes is the side-information of common parents but as binary value. In that case, the link-strength function would consider the additional term
\[\sum_k^{\ell^k_{ij}} v_m P_k(i,j),\]
where $P_k(i,j)$ is the parent adjacency matrix of the relation $k$ in the group structure defined as 
{\small
\begin{align}
P_k(i,j)= \left\{
\begin{array}{ll}
1 & \textrm{if $\mathbf{x}_i$ and $\mathbf{x}_j$ have common parents in relation $r_k$}\\
0 &\textrm{otherwise,}\\
\end{array}\right.
\end{align}}
\noindent and $\ell^k_{ij}$ is the number of common parents of $\mathbf{x}_i$ and $\mathbf{x}_j$ in the relation $r_k$.

With the link-strength function, we select the relative constraints set  $\mathcal{C}=\{ (i,j,k):LS(\mathbf{x}_i,\mathbf{x}_j ) \ge LS(\mathbf{x}_i,\mathbf{x}_k ) \} $ and use the constraints on two different classical metric learning algorithms, ITML  and LSML.

\section{Experiments}
\label{sec:exp}

\subsection{Datasets and Tasks}

We conduct experiments to compare the performance of the constraints generated by link-strength function and the constraints generated by the label information. To compare fairly, we set the amount of constraints generated by different ways are same and the formula are both in the relative distance constraint. Basically, any relational data for which classification is needed can be tackled by our proposition. 

We consider several real-world relational datasets which contains feature information for mapping with the learned metric, the target label information, and the relational information (existing links or valued links). We learn the metric of one entity table for predicting target label, so for the same dataset, we can learn different metrics for different tasks. Here are the descriptions of the chosen datasets and tasks: 

\begin{itemize}
\item Movie: MovieLens dataset \cite{harper2016movielens} is a classical relational dataset which is widely used in many related papers. It consists of a relational table which has 100,000 ratings (1-5) from 943 users on 1682 movies; a movie entity table with feature information about the movies; and a user entity table with id, age, gender, occupation, and other feature information on users. Each user has rated at least 20 movies so the relational supervised information is quite dense. We define two tasks on this dataset:
    \begin{itemize}
    \item Movie-item: We select the movies table as the entity table to learn the metric on. We choose the most popular genre as the target label and use the release date and other genres as the attributes.
    \item Movie-year: We select the users table as the entity table to learn the metric on. The age of users is discretized into 5 bins as the target label and the other feature information are the attributes. 
    \end{itemize}

\item BookCX: We also consider the book-crossing database \cite{ziegler2005improving}. We select a randomly sampled subset BookCX from the data. This subset contains 2,400 users giving 5,000 ratings (1-10) on 10,000 books. For this dataset, we use the bag-of-words model to encode the text information from the titles, the authors and the publishers into binary attributes. 
    \begin{itemize}
    \item BookCX-year: We consider the public year segmented into 5 bins as the target label and the bag-of-word of text information as the attributes. 
    \item BookCX-word: We apply PCA (Principal Component Analysis) on the bag-of-word of text information and limit the number of dimensions to 12. Then we randomly choose one of the processed dimension and segment it into 5 bins as the target label. The other features are considered as attributes.
    \end{itemize}

\item Citelike: There are two versions of citelike dataset, Citelike-t and Citelike-a, both used in the paper \cite{wang2013collaborative}. They were collected from CiteULike and Google Scholar. CiteULike allows users to create their own collections of articles. There are abstracts, titles, and tags for each article. They manually select hundreds of seed tags and collect all the articles with at least one of these tags. They also crawl the citations between the articles from Google Scholar. Note that the final number of tags associated with all the collected articles is far more than the number of seed tags. To reduce the computation complexity, we apply PCA on the large and sparse tag feature space and limit the number of dimensions to 12. We randomly choose one of the processed features and segment it into 5 bins as the target label. Note that the sampling of Citelike-t and Citelike-a are independent and the density of the links is different.

\item Mondial: This dataset \cite{may-MONDIAL-report-99} contains the relational version of the geographical Web data sources which is composed of CIA World Factbook, a predecessor of Global Statistics, additional textual sources for coordinates, the International Atlas and some geographical data of the Karlsruhe TERRA database. We chose part of the entity table City to learn and use the table Countries as the parent table. The population is segmented into 5 bins as the target label


\end{itemize}

\subsection{Result and Analysis}

For all the used datasets, the balance parameter between association attributes in the link-strength function is set to $\gamma = \frac{\alpha}{\alpha+\beta}$ in order to adapt to different situations of the datasets, where $\alpha$ is the number of numerical variables and $\beta$ is the number of categorical variables as mentioned before. 

The evaluation of the proposition is done by comparing the effect of learned metric with k-nearest-neighbour classification. For the set of k-nn classification, we use $k$ equal to 5 and score the performance with accuracy rate via randomly shuffled 3-fold cross validation. Note that we tried different values for $k$ (in particular 3, 5, 7 and 9), and the results were consistent with the results reported here for $k=5$ on most datasets. 

For each experiment, the number of constraints varies from 100 to 500, and we give the average value of each sets as the final result. All the experiments were run on a 3.1Ghz Intel Core i5 processor, with 16 Go 1867 MHz DDR3, and the code will be published for research reproducibility. We also give results obtained without learning a metric, i.e. using an Euclidean distance for k-nn algorithm (\texttt{Euc}).
Results are given in Tables \ref{tab:itml} and \ref{tab:lsml}.

\begin{table*}[t]
\centering
\begin{tabular}{c|ccccccc}
\hline
 ITML & Movie-item & Movie-user& BookCX-year & BookCX-word & Citelike-t & Citelike-a & Mondial \\ \hline
\texttt{Euc} & 98.58 $\pm$0.46 & 68.28 $\pm$4.00 & 36.16 $\pm$1.25 & 90.36 $\pm$0.91 & 88.03 $\pm$0.45 & 88.94 $\pm$0.55 & 68.97 $\pm$7.99 \\ 
 \texttt{Lab}  & 98.62 $\pm$0.52 & 67.72 $\pm$5.38 & 36.19 $\pm$1.09 & 89.57 $\pm$0.87 & 85.91 $\pm$0.62 & 89.89 $\pm$0.23 & 70.39 $\pm$8.29 \\ 
 \texttt{Rel}  & 97.48 $\pm$0.66 & 68.66 $\pm$3.40 & 36.12 $\pm$1.17 &  \textbf{91.29}  $\pm$0.74 & 90.76 $\pm$0.56 & 89.94 $\pm$0.41 & 71.55 $\pm$5.54 \\ 
  \texttt{Pro}  & 97.54 $\pm$0.42 & 69.04 $\pm$4.02 & 36.38 $\pm$1.62 & 90.33 $\pm$0.74 & 92.06 $\pm$0.46 &  \textbf{90.35}  $\pm$0.32 & 71.24 $\pm$7.77 \\ 
 \texttt{Both}  &   \textbf{98.67}  $\pm$0.50 &  \textbf{69.48} $\pm$3.08 & \textbf{36.97} $\pm$1.69 & 90.43 $\pm$0.72 &  \textbf{92.65} $\pm$0.40 & \textbf{90.35}  $\pm$0.32 & \textbf{72.00} $\pm$7.16 \\ \hline
\end{tabular}
\caption{The accuracy score of knn with ITML}
\label{tab:itml}
\end{table*}

\begin{table*}[t]
\centering
\begin{tabular}{c|ccccccc}
\hline
 LSML  & Movie-item & Movie-user& BookCX-year & BookCX-word & Citelike-t & Citelike-a & Mondial \\ \hline
 \texttt{Euc} & 98.58 $\pm$0.46 & 68.28 $\pm$4.00 & 36.16 $\pm$1.25 & 90.36 $\pm$0.91 & 88.03 $\pm$0.45 & 88.94 $\pm$0.55 & 68.97 $\pm$7.99 \\
 \texttt{Lab}  & 99.06 $\pm$0.58 & 65.92 $\pm$5.38 & 36.36 $\pm$1.06 & 94.46 $\pm$1.06 & 85.53 $\pm$0.67 & 94.62 $\pm$0.33 & 68.45 $\pm$8.47 \\ 
 \texttt{Rel}  & 98.63 $\pm$0.52 & 66.67 $\pm$4.03 & 36.21 $\pm$1.21 & 94.91 $\pm$0.41 & 85.65 $\pm$0.63 & 94.62 $\pm$0.65 & 70.25 $\pm$6.47 \\ 
 \texttt{Pro}  & 98.63 $\pm$0.40 &  \textbf{66.98}  $\pm$4.63 & \textbf{36.42} $\pm$1.52 &  \textbf{94.92}  $\pm$0.61 &  \textbf{85.69}  $\pm$0.57 &  \textbf{94.63}  $\pm$0.42 & 70.62 $\pm$7.04 \\ 
 \texttt{Both}  &   \textbf{99.12}  $\pm$0.52 &  \textbf{66.98}  $\pm$4.63 & \textbf{36.42 $\pm$1.15} &  \textbf{94.92}  $\pm$0.61 &  \textbf{85.69}  $\pm$0.57 &  \textbf{94.63 $\pm$ 0.30} & \textbf{71.03} $\pm$7.35 \\ \hline \hline
\end{tabular}
\caption{The accuracy score of knn with LSML}
\label{tab:lsml}
\end{table*}
In Tables \ref{tab:itml} and \ref{tab:lsml}, \texttt{Lab} indicates the result obtained using only the constraints generated from label, \texttt{Rel} shows the result obtained by the constraints generated from the relative link constraints, i.e. using the adjacency matrix $A$ of the graph. \texttt{Pro} gives the performance of our proposition based on link-strength constraints and \texttt{Both} shows the best result with both label constraints and the link-strength constraints while the proportion of them are appropriated. The Table \ref{tab:itml_2} and \ref{tab:lsml_2} show the results with different set of proportion of label constraints and the link-strength constraints. Proportion equal to 1 corresponds to the situation of using only labels, and a proportion of 0 corresponds to the fact of using only link-strength based constraints. As can be seen in the Table, results tend to be better when using mostly link-strength constraints.

\begin{table*}[t!]
\centering
\begin{tabular}{c|ccccccc}
\hline
   Proportion & Movie-item & Movie-user& BookCX-year & BookCX-word & Citelike-t & Citelike-a & Mondial \\ \hline
 \texttt{1.0} & 98.62  $\pm$0.52 & 67.72  $\pm$5.38 & 36.19  $\pm$1.09 & 89.57  $\pm$0.87 & 85.91  $\pm$0.62 & 89.89  $\pm$0.23 & 70.39  $\pm$8.29 \\ 
 \texttt{0.8}  & 98.52  $\pm$0.23 & 66.21  $\pm$3.71 & 36.21  $\pm$1.06 & 89.52  $\pm$0.55 & 86.71  $\pm$0.67 & 89.71  $\pm$0.42 & \textbf{72.00}  $\pm$7.16 \\ 
 \texttt{0.6}  & 98.42  $\pm$0.54 & 66.87  $\pm$4.09 & 36.66  $\pm$1.67 & 89.41  $\pm$0.82 &  \textbf{92.65}  $\pm$0.40 & 89.61  $\pm$0.31 & 71.03  $\pm$6.22 \\ 
 \texttt{0.4} &   \textbf{98.67}   $\pm$0.50 & 67.65  $\pm$4.82 & \textbf{36.97}  $\pm$1.69 & 88.87  $\pm$0.78 & 90.87  $\pm$0.38 & 89.91  $\pm$0.35 & 70.82  $\pm$7.66 \\ 
 \texttt{0.2} & 97.32  $\pm$1.52 &  \textbf{69.48}  $\pm$3.08 & 36.28  $\pm$1.16 &  \textbf{90.43}  $\pm$0.72 & 91.24  $\pm$0.41 & 90.21  $\pm$0.37 & 69.21  $\pm$7.98 \\ 
 \texttt{0.0} & 97.54  $\pm$0.42 & 69.04  $\pm$4.02 & 36.38  $\pm$1.62 & 90.33  $\pm$0.74 & 92.06  $\pm$0.46 &  \textbf{90.35}   $\pm$0.32 & 71.24  $\pm$7.77 \\  \hline
\end{tabular}
\caption{The accuracy score of knn with ITML while the proportion of label constraints and the link-strength constraints gradient change from full label constraints to full link-strength constraints. }
\label{tab:itml_2}
\end{table*}

\begin{table*}[t!]
\centering
\begin{tabular}{c|ccccccc}
\hline
   Proportion  & Movie-item & Movie-user& BookCX-year & BookCX-word & Citelike-t & Citelike-a & Mondial \\ \hline
 \texttt{1.0}  & 99.06  $\pm$0.58 & 65.92  $\pm$5.38 & 36.36  $\pm$1.06 & 94.46  $\pm$1.06 & 85.53  $\pm$0.67 & 94.62  $\pm$0.33 & 68.45  $\pm$8.47 \\ 
 \texttt{0.8}  & 99.04  $\pm$0.54 & 65.67  $\pm$4.32 & 36.22  $\pm$1.14 & 94.89  $\pm$0.60 & 85.46  $\pm$0.61 & 94.59  $\pm$0.38 & \textbf{71.03}  $\pm$7.35 \\ 
 \texttt{0.6}  &   \textbf{99.12}   $\pm$0.52 & 65.78  $\pm$3.80 & 36.13  $\pm$1.18 & 94.78  $\pm$0.58 & 85.53  $\pm$0.79 & 94.61  $\pm$0.41 & 70.22  $\pm$8.42 \\ 
 \texttt{0.4} & 99.08  $\pm$0.67 & 66.21  $\pm$4.52 & 36.3  $\pm$1.50 & 94.85  $\pm$0.68 & 85.45  $\pm$0.70 &  \textbf{94.63 $\pm$0.30} & 69.79  $\pm$6.54 \\ 
 \texttt{0.2} & 98.87  $\pm$0.62 & 66.14  $\pm$5.39 & \textbf{36.42 $\pm$1.15} & 94.83  $\pm$0.86 & 85.45  $\pm$0.67 &  \textbf{94.63}   $\pm$0.52 & 70.67  $\pm$7.22 \\ 
 \texttt{0.0} & 98.63  $\pm$0.40 &  \textbf{66.98}   $\pm$4.63 & \textbf{36.42}  $\pm$1.52 &  \textbf{94.92}   $\pm$0.61 &  \textbf{85.69}   $\pm$0.57 &  \textbf{94.63}   $\pm$0.42 & 70.62  $\pm$7.04 \\ \hline
\end{tabular}
\caption{The accuracy score of knn with LSML while the proportion of label constraints and the link-strength constraints gradient change from full label constraints  to full link-strength constraints. }
\label{tab:lsml_2}
\end{table*}

As can be seen, except on Movie-item task and BookCX-year task, comparing with constraints generated only from labels, the link-strength constraints lead to a great improvement of accuracy. On most datasets, the link-strength constraints shows better performance than the relative link constraints, except the BookCX-word task with ITML. For both labels and relational information, it provides a better accuracy than the constraints obtained from labels and similar to the constraints generated with link-strength function. Considering the different number of references in these datasets, for example 100, 000 references for 943 entities for Movie-user and 5,000 references for 10,000 entities for BookCX-year, we speculate that density or sparsity of the references leads to the deviation of results.

\section{Conclusion}
\label{sec:con}
In this paper, we propose a simple, yet effective, way of learning a metric dedicated to (multi-)relational data.
This work on relational metric learning clearly shows the benefit, in terms of accuracy, of considering relational information between entities instead of the sole consideration of labels. As a first perspective, we plan to consider other way of computing link strength, that may be inspired from graph analysis techniques, e.g. connection strength metric, length of the shortest path, value of the maximum network flow between nodes. In particular, we want to consider slot chains (i.e. sequences of foreign key references) which are longer than 1. We also plan to define a dedicated relational metric that could be learned directly, instead of setting relational constraints on standard metric learning algorithms. 

\bibliographystyle{aaai}
\bibliography{ref}

\end{document}